\documentclass[letterpaper]{article} % DO NOT CHANGE THIS
\usepackage[submission]{aaai2027}  % DO NOT CHANGE THIS
\usepackage[hyphens]{url}  % DO NOT CHANGE THIS
\usepackage{graphicx} % DO NOT CHANGE THIS
\usepackage{natbib}  % DO NOT CHANGE THIS AND DO NOT ADD ANY OPTIONS TO IT
\usepackage{caption} % DO NOT CHANGE THIS AND DO NOT ADD ANY OPTIONS TO IT
\usepackage{algorithm}
\usepackage{algorithmic}

\usepackage{newfloat}
\usepackage{listings}
\DeclareCaptionStyle{ruled}{labelfont=normalfont,labelsep=colon,strut=off} % DO NOT CHANGE THIS
\floatstyle{ruled}
\newfloat{listing}{tb}{lst}{}
\floatname{listing}{Listing}

\usepackage{booktabs}

\usepackage{amsmath}
\usepackage{amssymb}

\title{Removing Temporal Note Redundancy Improves Multimodal Reinforcement Learning for Medicine}
\author{
Chenran Weng\textsuperscript{\rm 1},
Joo Seung Lee\textsuperscript{\rm 1},
Malini Mahendra\textsuperscript{\rm 2},
Anil Aswani\textsuperscript{\rm 1}\corresponding
}

\affiliations{
\textsuperscript{\rm 1}University of California, Berkeley\\
Berkeley, CA, USA\\
\textsuperscript{\rm 2}University of California, San Francisco\\
San Francisco, CA, USA\\
chenrann@berkeley.edu, jooseung\_lee@berkeley.edu,
malini.mahendra@ucsf.edu, aaswani@berkeley.edu
}

\begin{document}

\maketitle

\begin{abstract}
Mechanical ventilation is a critical life-support intervention, requiring dynamic adjustments to ventilator settings as a patient's condition evolves. While reinforcement learning (RL) offers a promising framework for optimizing these sequential decisions, standard approaches rely primarily on structured electronic health record (EHR) data, missing crucial clinical context recorded in free-text notes. Integrating longitudinal clinical notes into RL state spaces is challenging because notes are heavily inflated by temporal redundancy, such as copy-forward text, templating, and repetitive documentation, which dilutes time-local updates and degrades state representation quality. To address this, we propose a redundancy-aware multimodal state representation framework that explicitly removes duplicated note text over time before policy learning. We evaluate two computationally efficient temporal decomposition strategies for removing duplicated note text: (1) an embedding-space decomposition using singular value decomposition on local history subspaces, and (2) an interpretable sentence-level diff operation that filters out previously documented sentences before text encoding. Using real-world ICU data, we demonstrate that state representations constructed by stripping temporal note redundancy significantly outperform both structured-only and raw-note baselines across multiple off-policy evaluation methods (Model-Based Rollouts, Fitted Q-Evaluation, Weighted Importance Sampling, and Weighted Doubly Robust Evaluation). Our findings show that explicitly isolating new clinical information from repeated note text yields higher-quality state representations and directly improves RL performance for clinical decision support.
%Mechanical ventilation is a key form of life support for patients with impaired respiratory function, requiring clinicians to repeatedly adjust ventilator settings as each patient’s condition evolves. Offline reinforcement learning provides a promising framework for learning sequential treatment policies from retrospective ICU trajectories, but most existing approaches rely mainly on structured electronic health record variables and may miss clinical context documented in free-text notes. Although clinical notes can provide complementary information for ventilator decision-making, repeated documentation, templating, and copy-forward practices can make their representations heavily influenced by redundant content. In this paper, we study redundancy-aware multimodal state representation for offline reinforcement learning in mechanical ventilation. We propose two approaches that decompose clinical notes into history-related and new-information components. Using ICU data from MIMIC-III, we show that the resulting state representations outperform both structured-only states and raw-note states across multiple off-policy evaluation methods. These results suggest that clinical notes can improve offline reinforcement learning for healthcare decision support when their temporal redundancy is explicitly addressed.
\end{abstract}

\section{Introduction}
Mechanical ventilation is an intervention in intensive care for patients unable to maintain adequate respiratory function. By delivering controlled air and oxygen, it assists or replaces spontaneous breathing. Effective management requires clinicians to continuously adjust ventilator settings as patient conditions evolve, while inappropriate settings can cause complications like ventilator-induced lung injury, diaphragm dysfunction, pneumonia, and oxygen toxicity \cite{pham2017mechanical}. These risks make mechanical ventilation an important sequential decision-making problem.

Offline reinforcement learning provides a natural framework for learning time-varying treatment strategies from retrospective ICU trajectories. Unlike supervised learning, which primarily imitates clinician actions, RL aims to optimize long-term outcomes from sequential states, actions, and rewards. Prior studies have applied RL to mechanical ventilation using ICU data \cite{peine2021development,kondrup2023towards,lee2025matching,den2024guideline}. However, these approaches mainly define patient states using structured EHR variables, which may not fully capture the clinical context considered during ventilator decisions.

Clinical notes provide a natural complementary source of such context because they contain clinician assessments, treatment plans, and descriptions of patient status. However, using longitudinal notes directly is challenging because clinical documentation is often highly redundant over time due to templating, repeated documentation, copy-paste, and copy-forward practices. In a study of approximately 2.3 million inpatient notes, Vawdrey et al. found that 42\% of inpatient notes used copy-paste and 19\% of inpatient note content was copied \cite{vawdrey2022practical}. For offline RL, this creates a state representation problem. A raw note embedding may mix persistent historical content, repeated text, and newly introduced clinical updates, making decision-relevant changes harder to identify.

Although modern language representation methods like large language models can provide a potential approach for summarizing or removing redundant information from clinical notes, applying them repeatedly to longitudinal note histories can be computationally and financially expensive due to the large number of patients and decision times in ICU trajectories. This motivates the need for a more efficient approach that can identify temporal redundancy while preserving clinically meaningful information for sequential decision-making. 

To address this challenge, we make the following contributions in this paper:

\begin{enumerate}
    \item We introduce two computationally efficient note decomposition methods, one in embedding space and one at the sentence level, to separate history-related content from newly introduced information without relying on expensive large language model processing. These are our proposed methods for removing temporal note redundancy in a computationally efficient way.
    
    \item We formulate a redundancy-aware multimodal state construction framework for offline RL in mechanical ventilation.
    
    \item We evaluate the resulting state representations on MIMIC-III mechanical ventilation trajectories and show that they improve estimated policy value over structured-only and raw-note multimodal baselines across multiple off-policy evaluation methods. This supports the accuracy and efficacy of our approach.
\end{enumerate}

\section{Related Work}
\subsection{Mechanical Ventilation Optimization via RL}
RL has been increasingly studied for mechanical ventilation optimization, with prior work addressing dynamic ventilator control, model-based policy learning, interpretability, safety, and weaning. Peine et al. introduced VentAI, a tabular Q-learning framework that uses structured patient trajectories to recommend ventilation regimes and evaluate their association with mortality \cite{peine2021development}. Beyond this data-driven control formulation, Chen et al. explored a model-based hybrid soft actor-critic approach for learning ventilator settings \cite{chen2022model}, while Lee et al. emphasized interpretability and transparent off-policy evaluation for making learned ventilation policies more clinically meaningful \cite{lee2025matching}. Another line of work focuses on safety and reward design: DeepVent applies conservative offline RL with clinically relevant intermediate rewards \cite{kondrup2023towards}, Zhang et al. balance therapeutic effect and safety in ventilator recommendation \cite{zhang2024balancing}, Eghbali et al. introduce uncertainty-aware recommendations through conformal deep Q-learning \cite{eghbali2025distribution}, and den Hengst et al. incorporate clinical guidelines into the RL learning signal and safety constraints \cite{den2024guideline}. Related studies also consider mechanical ventilation weaning and extubation readiness using off-policy RL and inverse RL \cite{prasad2017reinforcement,yu2019inverse}.

\subsection{Multimodal EHR Representation Learning}

Multimodal EHR learning has shown that clinical notes can complement structured variables for medical prediction. Prior work fuses time-invariant features, time-series measurements, and clinical notes \cite{yang2021leverage}, and further shows that the gain from notes can depend strongly on the clinical information they contain rather than only on model architecture \cite{husmann2022importance}. Recent studies extend structured-text fusion to ICU outcome prediction \cite{lyu2023multimodal, ruan2025evidence}, and temporal modeling, including early sepsis prediction with physiological time series and ClinicalBERT note features \cite{wang2022integrating}, and cross-modal temporal pattern discovery in EHR trajectories \cite{wang2025ctpd}. Broader multimodal EHR pretraining methods also aim to learn general representations across tasks \cite{wang2023hierarchical}. Closest to our setting, MORE-CLEAR incorporates clinical notes into offline RL state representations for sepsis treatment \cite{lim2025more}. However, these studies generally treat note representations as additional context, rather than asking how longitudinal note redundancy should be decomposed when notes become part of an RL state.

\subsection{Redundancy and Copy-Forward in Clinical Notes}

Clinical notes often contain repeated information because of templates, copy-paste, and copy-forward practices. Prior work has measured redundancy in EHR corpora and studied how it affects text mining performance \cite{cohen2013redundancy}. Other studies have described where duplicate information in electronic medical records comes from \cite{steinkamp2022prevalence} and proposed practical tools for monitoring copy-paste use in clinical notes \cite{vawdrey2022practical}. This repeated content can have real effects: limiting copy-paste has been linked to better inpatient care quality \cite{cheng2022restricted}, note bloat can affect deep learning models for clinical prediction \cite{liu2022note}, and recent work removes repeated chart text to make LLM-based clinical systems more efficient while keeping useful clinical information \cite{cahoon2026clinical}. These studies motivate redundancy reduction, but they do not address how repeated and newly introduced note content should be separated for sequential decision-making states; our work studies that gap in offline RL for mechanical ventilation.

\section{Method}
\label{sec:method}

\subsection{Mechanical Ventilation MDP}
\label{subsec:mv_mdp}

We formulate mechanical ventilation management as a finite-horizon
offline RL problem. Each patient trajectory is
represented as a sequence of transitions
$(s_t,a_t,r_t,s_{t+1})_{t=1}^{T},$
where $s_t$ denotes the patient state at time $t$, $a_t$ denotes the
ventilator action, and $r_t$ is the clinical reward. Following prior
work on RL for mechanical ventilation, we use
4-hour decision intervals and define the structured MDP using the
same observable clinical variables, action discretization, and reward
function as in \cite{lee2025matching}. This design isolate the effect of the proposed state representations while
keeping the clinical decision problem fixed.

The structured component of the state includes the following variables:
\begin{itemize}
    \item Respiratory: Respiratory rate, Spontaneous tidal volume,
PaO2FiO2 Ratio, Mean airway pressure
    \item Hemodynamic: Heart rate, Systolic BP, Diastolic BP
    \item Blood Gas: SpO2, PaCO2, PaO2
    \item Miscellaneous: Sepsis, Weight, Age, GCS, Cumulative Fluid
Balance
\end{itemize}
Following the original MDP formulation, these observable variables are also
augmented with a time-invariant propensity score estimated from additional patient-type features using logistic regression. In our experiments, this structured state serves as the baseline representation, to which different clinical-note representations are added.

\paragraph{Action space.}
At each decision time, the action corresponds to a discretized
ventilator setting. Following the same action definition as the prior
work, the action is a tuple
$a_t = (a_t^{\mathrm{Vt}}, a_t^{\mathrm{PEEP}}, a_t^{\mathrm{FiO_2}}),$
% \[
% a_t = (a_t^{\mathrm{Vt}}, a_t^{\mathrm{PEEP}}, a_t^{\mathrm{FiO_2}}),
% \]
where $a_t^{\mathrm{Vt}}$ denotes ideal-body-weight-adjusted tidal
volume, $a_t^{\mathrm{PEEP}}$ denotes positive end-expiratory pressure,
and $a_t^{\mathrm{FiO_2}}$ denotes fraction of inspired oxygen.
Although these ventilator settings are continuous in clinical
practice, they are discretized into clinically meaningful bins through clustering analysis. This
yields a finite discrete action space suitable for offline discrete
RL.

\paragraph{Reward function.}
We use the same clinically motivated reward function as prior work.
The reward encourages improvement in blood oxygen saturation while
penalizing aggressive ventilator settings. Specifically, the oxygenation
component rewards increases in SpO$_2$ only when both the current
and next SpO$_2$ values are below 95\%, avoiding additional reward
for increasing oxygen saturation beyond a clinically safe range:
\begin{equation}
    r_{\mathrm{SpO}_2}(s_t,s_{t+1}) = \begin{cases} s_{t+1}(\text{SpO}_{\text{2}}) - s_{t}(\text{SpO}_{\text{2}})\\ \quad\text{if $s_{t+1}$(SpO$_2$)$< 95 \wedge s_{t}$(SpO$_2$) $<95$} \\ 0 \qquad\text{otherwise}\end{cases}
\end{equation}
The action penalty discourages high tidal volume and high inspired
oxygen: $r_a(a_t)
=
-\alpha \mathbf{1}\{a_t^{\mathrm{Vt}} \geq 10\}
-\beta \mathbf{1}\{a_t^{\mathrm{FiO_2}} \geq 0.6\}$. The total reward is $r(s_t,a_t,s_{t+1})
=
r_{\mathrm{SpO}_2}(s_t,s_{t+1}) + r_a(a_t)$. We use this reward specification to maintain the same clinical
objective across all state representations. Therefore, differences
in policy performance can be attributed primarily to the proposed
state representation rather than to changes in the action space or
reward design.

\subsection{Problem Setup}
\label{subsec:problem_setup}
We study offline RL from ICU trajectories. At each time step \(t\), the observed patient state contains structured clinical variables \(u_t \in \mathbb{R}^{d_s}\) and clinical note text \(D_t\), yielding transitions \(\mathcal{D} = \{(u_t, D_t, a_t, r_t, u_{t+1}, D_{t+1})\}\) with action \(a_t \in \mathcal{A}\), reward \(r_t\), and next state \(s_{t+1}\). To avoid temporal leakage, \(D_t\) contains only notes available up to the state-construction cutoff for that 4-hour interval; later notes are not used.

A central challenge is that clinical notes contain substantial redundancy due to copy-forward, templating, and repeated historical content. As a result, a raw note embedding may mix together persistent background information and newly introduced clinical information, which can obscure the time-local patient state relevant for control. Our goal is therefore to construct a redundancy-aware note representation and fuse it with the structured variables for downstream offline RL.

Our approach is based on the same principle across all variants: a note should be decomposed into a history component and an innovation component before it is used as part of the RL state. We consider two instantiations of this idea.

First, we introduce an \emph{embedding-space temporal decomposition}, where the current note embedding is projected onto a patient-local subspace spanned by previous notes, and the residual is treated as the innovation component.

Second, we introduce a \emph{sentence-level diff decomposition}, where we first compare the current note against previous notes at the sentence level, explicitly split the note text into historical and newly added sentences, and then embed these two text segments separately.

The final RL state is formed by concatenating the structured variables with the compressed note-derived history and new-information features.

\begin{figure*}[t]
\centering
\includegraphics[width=0.8\textwidth]{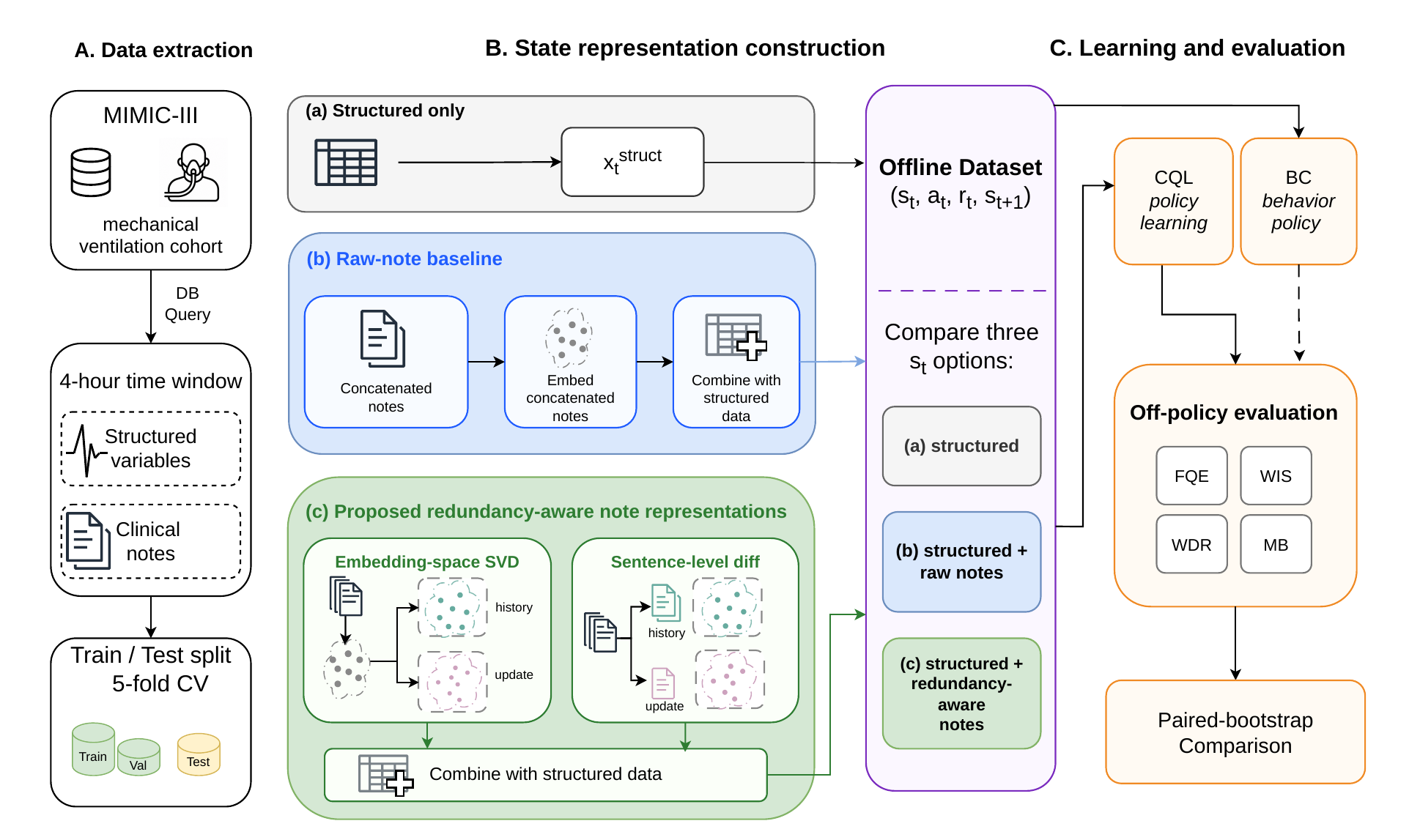}
\caption{Overview of the redundancy-aware multimodal state construction and evaluation pipeline.}
\label{fig:pipeline}
\end{figure*}

Figure~\ref{fig:pipeline} summarizes the proposed pipeline. Starting from the MIMIC-III mechanical ventilation cohort, we align structured variables and clinical notes to the same 4-hour decision intervals. We then construct three state representations for comparison: a structured-only state, a raw-note multimodal baseline, and the proposed redundancy-aware multimodal states. The redundancy-aware representations decompose clinical notes into history-related and update-related components before combining them with structured clinical variables. The resulting states are used with the same offline policy-learning and off-policy evaluation pipeline, allowing us to isolate the effect of state representation.

\subsection{Note Encoding}
\label{subsec:note_encoding}

For each note \(D_t\), we compute $x_t=f(D_t)\in\mathbb{R}^{d_n}$ using ClinicalBERT \cite{alsentzer2019publicly}. Intervals with no note text use a zero note vector. The same pretrained encoder is used for the raw-note baseline, the embedding-space decomposition, and the sentence-level diff decomposition, keeping the comparison focused on the redundancy-handling strategy rather than on the text encoder.

% For each note \(D_t\), we compute a representation using ClinicalBERT\cite{alsentzer2019publicly}. Let
% \[
% f: \mathcal{T} \rightarrow \mathbb{R}^{d_n}
% \]
% denote the ClinicalBERT encoder, where \(\mathcal{T}\) is the space of note texts. We define the note embedding as
% \[
% x_t = f(D_t) \in \mathbb{R}^{d_n}.
% \]
% In our implementation, ClinicalBERT is used as the note encoder for all note-based state representations. Thus, the same pretrained clinical language model is used for the raw-note baseline, the embedding-space decomposition, and the sentence-level diff decomposition. This keeps the comparison focused on the redundancy-handling strategy rather than on differences in the underlying text encoder.

\subsection{Embedding-Space Temporal Decomposition}
\label{subsec:svd_decomposition}

The first representation variant performs deduplication after note embedding. The main idea is to characterize the subspace of note content already explained by recent note history and then extract the residual as the current innovation.

Fix a time $t$, and consider the previous $k$ notes from the same ICU stay: $x_{t-k}, x_{t-k+1}, \dots, x_{t-1}.$  For early time steps $t<k$, we use the available history only; if the history is empty, the history component is set to zero and \(x_t\) is treated as new information.
Define the local history mean $\mu_t^{\mathrm{hist}} = \frac{1}{k}\sum_{i=t-k}^{t-1} x_i$. We then construct the centered history matrix
\begin{equation*}
    H_t =
\begin{bmatrix}
(x_{t-k} - \mu_t^{\mathrm{hist}})^\top \\
(x_{t-k+1} - \mu_t^{\mathrm{hist}})^\top \\
\vdots \\
(x_{t-1} - \mu_t^{\mathrm{hist}})^\top
\end{bmatrix}
\in \mathbb{R}^{k \times d_n}.
\end{equation*}
We compute
\(H_t=U_t\Sigma_tV_t^\top\), and let \(V_{t,\ell}\) contain the top \(\ell\)
right singular vectors, which span the recent-history subspace. With
\(y_t=x_t-\mu_t^{\mathrm{hist}}\), the projection and residual are $x_t^{\mathrm{svd,hist}}=\mu_t^{\mathrm{hist}}+V_{t,\ell}V_{t,\ell}^{\top}y_t$ and $x_t^{\mathrm{svd,new}}=(I-V_{t,\ell}V_{t,\ell}^{\top})y_t$. The original note embedding then admits the additive decomposition \(x_t = x_t^{\mathrm{svd,hist}} + x_t^{\mathrm{svd,new}}\).
Here $x_t^{\mathrm{svd,hist}}$ represents the part of the current note embedding explained by recent note history, while $x_t^{\mathrm{svd,new}}$ captures the information not explained by the local history subspace.

\subsection{Sentence-Level Diff Decomposition}
\label{subsec:diff_decomposition}

The second representation variant performs deduplication before note embedding. The main idea is to explicitly separate the current note text into sentences that already appeared in the recent note history and sentences that are newly introduced at the current time step.

Let $\mathcal{S}_t = (c_{t,1}, c_{t,2}, \dots, c_{t,m_t})$
denote the ordered sentence sequence of note $D_t$, where $c_{t,j}$ is the $j$-th sentence in the note. Let the sentence history over the previous $k$ notes be $\mathcal{R}_t = \mathcal{S}_{t-k} \oplus \mathcal{S}_{t-k+1} \oplus \cdots \oplus \mathcal{S}_{t-1}$, where $\oplus$ denotes temporal concatenation of sentence sequences. 
%Here, each sentence is treated as one sequence element, analogous to treating each line as one element in standard text differencing.
We apply a sentence-level diff operator based on the classical
longest-common-subsequence formulation of differential file comparison
\cite{hunt1976algorithm}:
$(\mathcal{S}_t^{\mathrm{hist}}, \mathcal{S}_t^{\mathrm{new}})
=
\operatorname{Diff}(\mathcal{S}_t, \mathcal{R}_t)$, where $\mathcal{S}_t^{\mathrm{hist}}$ contains the current-note sentences aligned to the history sequence and
$\mathcal{S}_t^{\mathrm{new}}$ contains the current-note sentences not aligned to the history sequence.

We reconstruct two note texts: $D_t^{\mathrm{hist}} = \operatorname{concat}(\mathcal{S}_t^{\mathrm{hist}})$ and $D_t^{\mathrm{new}} = \operatorname{concat}(\mathcal{S}_t^{\mathrm{new}})$, where $\operatorname{concat}(\cdot)$ concatenates a sentence sequence into text. We then embed the two views separately: $x_t^{\mathrm{diff,hist}} = f(D_t^{\mathrm{hist}})$ and $x_t^{\mathrm{diff,new}} = f(D_t^{\mathrm{new}})$. If either sentence set is empty, we use the zero vector for the corresponding embedding.

Compared with the embedding-space decomposition, the sentence-level diff decomposition removes redundancy earlier in the pipeline and does so in a more explicit and interpretable way. The ``history'' channel contains only text that can be traced to previous notes, while the ``new'' channel contains only text newly introduced at time \(t\). This representation is particularly natural when note duplication occurs via sentence-level copy-forward.

\subsection{Compression of Note Components}
\label{subsec:compression}
All note components remain high-dimensional and are therefore compressed before being used in downstream RL. We apply PCA fitted on the training set to each component separately, producing low-dimensional history and new-information features for both the embedding-space and sentence-level decompositions. This step controls state dimension and is not essential to the core redundancy decomposition.

For the embedding-space decomposition, we obtain low-dimensional features $z_t^{\mathrm{svd,hist}} \in \mathbb{R}^{q_{\mathrm{svd,hist}}}$ and $z_t^{\mathrm{svd,new}} \in \mathbb{R}^{q_{\mathrm{svd,new}}}$ from $x_t^{\mathrm{svd,hist}}$ and $x_t^{\mathrm{svd,new}}$, respectively.

For the sentence-level diff decomposition, we similarly obtain
$z_t^{\mathrm{diff,hist}} \in \mathbb{R}^{q_{\mathrm{diff,hist}}}$ and $z_t^{\mathrm{diff,new}} \in \mathbb{R}^{q_{\mathrm{diff,new}}}$ from $x_t^{\mathrm{diff,hist}}$ and $x_t^{\mathrm{diff,new}}$.

We summarize the different state constructions in Table~\ref{tab:state_repr}. Both variants provide the RL model with two note-derived channels: a history-like channel and a newly introduced information channel. They differ in the stage at which redundancy is addressed. The embedding-space method performs decomposition after embedding, modeling redundancy as low-rank structure in the continuous representation space, whereas the sentence-level diff method removes redundancy before embedding by identifying explicit textual reuse. These approaches are complementary: the former is naturally aligned with continuous representation learning, while the latter provides an interpretable text-level decomposition for sentence-level copy-forward. In both cases, the goal is to construct note-based state representations that are less dominated by redundant content and more responsive to changes in the patient’s clinical condition.

\begin{table}[t]
\centering
%\small
\begin{tabular}{ll}
\toprule
\textbf{Method} & \textbf{State Representation} \\
\midrule
Structured-only 
& $s_t^{\mathrm{struct}} = u_t$ \\

Structured + raw note 
& $s_t^{\mathrm{raw}} = [u_t, z_t^{\mathrm{raw}}]$ \\

Embedding-space SVD 
& $s_t^{\mathrm{svd}} = [u_t, z_t^{\mathrm{svd,hist}}, z_t^{\mathrm{svd,new}}]$ \\

Sentence-level diff 
& $s_t^{\mathrm{diff}} = [u_t, z_t^{\mathrm{diff,hist}}, z_t^{\mathrm{diff,new}}]$ \\ \\
\bottomrule
\end{tabular}
\caption{Summary of state representations.}
\label{tab:state_repr}
\end{table}

\subsection{Offline Policy Learning}
\label{subsec:offline_policy_learning}
For each state representation in Table~\ref{tab:state_repr}, we construct transitions \((s_t,a_t,r_t,s_{t+1})\) and train policies on the same training trajectories. This ensures that all downstream comparisons use the same patient population, action space, reward function, and policy-learning algorithm.

\paragraph{Behavior cloning.}
We train a behavior cloning (BC) policy \cite{bain1995framework} to imitate historical clinician
actions by solving the supervised classification problem. 
% $\hat{\pi}_{\mathrm{BC}}
% =
% \arg\min_{\pi}
% \sum_{(s_t,a_t)\in\mathcal{D}_{\mathrm{train}}}
% -\log \pi(a_t \mid s_t)$.
BC serves as a clinician-imitation baseline and is used to estimate the behavior policy for importance-ratio-based off-policy evaluation.

\paragraph{Conservative Q-learning.}
Our main offline RL policy is trained using Conservative Q-Learning (CQL)
\cite{kumar2020conservative}, which mitigates overestimation and
out-of-distribution action selection by learning a conservative Q-function.
For each state representation, we train a separate CQL policy
$\hat{\pi}_{\mathrm{CQL}}(s)=\arg\max_{a\in\mathcal{A}}Q_\theta(s,a).$
All policies use the same CQL algorithm and training procedure, with the
only difference being the underlying state representation.
\section{Experiments}

\subsection{Dataset}
This study uses data from MIMIC-III, a large-scale and openly accessible
critical care database~\cite{johnson2016mimic}. We focus on ICU stays involving
mechanical ventilation and follow the cohort construction and structured-data
preprocessing pipeline of~\cite{lee2025matching}. In brief, patient trajectories are
discretized into 4-hour intervals, and each episode contains demographic
variables, laboratory measurements, vital signs, and ventilator-related
variables. We include adult patients with documented 90-day mortality outcomes
and recorded tidal volume measurements. 

In pre-processing procedure, continuous structured variables are winsorized to reduce
the effect of extreme outliers, missing values are filled using a combination
of forward/backward filling and KNN imputation, and trajectories
are truncated to the first 72 hours of ventilation. This gives a maximum of 18
decision points per episode, with one decision point every 4 hours.

In addition to the structured variables, we extract clinical notes associated
with each ICU stay. Notes are aligned to the same 4-hour decision intervals as the structured observations. For each interval, we concatenate all notes whose timestamps fall within the interval, preserving the de-identified note time and
note category as headings. The final note dataset contains 157,924 notes from 10,125 ICU stays, aligned to 147,997 decision intervals. Each ICU stay contains 15.60 notes on average.

\subsection{Off-Policy Evaluation}
We evaluate each learned target policy $\pi_e$ using four complementary off-policy estimators: (i) model-based rollouts (MB), (ii) Fitted Q Evaluation (FQE), (iii) Weighted Importance Sampling (WIS), and (iv) Weighted Doubly Robust evaluation (WDR). Each estimator carries different bias--variance tradeoffs, and we report all four to triangulate policy value. Throughout, $\gamma$ denotes the discount factor and a trajectory $\tau = (s_0, a_0, r_0, s_1, a_1, r_1, \ldots, s_T)$ has length $T$.
\subsubsection{Model-Based Evaluation}
We learn a probabilistic transition model $\hat{P}(s' \mid s, a)$ from the offline dataset, and evaluate $\pi_e$ by Monte Carlo rollouts in the learned model. Starting from each test-set initial state $s_0$, we sample a trajectory $\hat\tau = (s_0, \hat a_0, \hat s_1, \hat a_1, \ldots)$ where $\hat a_t \sim \pi_e(\cdot \mid \hat s_t)$ and $\hat s_{t+1} \sim \hat P(\cdot \mid \hat s_t, \hat a_t)$, and accumulate the discounted reward
\begin{equation}
\textstyle\hat V^{\text{MB}}(\pi_e) \;=\; \mathbb{E}_{s_0 \sim \mathcal{D}_0}\left[ \sum_{t=0}^{T-1} \gamma^t \, \hat r(\hat s_t, \hat a_t) \right],
\end{equation}
where $\hat r(\cdot, \cdot)$ is the same reward function used during policy training and $\mathcal{D}_0$ is the empirical distribution of test-set initial states. To reduce variance from stochastic rollouts, we average over 5 independent trajectories per initial state. MB is unbiased only if the learned $\hat P$ matches the true transition kernel; in practice it can be optimistic when $\pi_e$ visits state--action regions with sparse training support.

\subsubsection{Fitted Q Evaluation}
Fitted Q Evaluation (FQE) \citep{le2019batch} estimates policy value by
iteratively fitting a Q-function under the evaluation policy. The final value
is computed from the learned Q-function:
\begin{equation}
\hat V_{\mathrm{FQE}}(\pi_e)
=
\mathbb{E}_{s_0\sim\mathcal{D}_0}
\left[
\mathbb{E}_{a_0\sim\pi_e(\cdot|s_0)}
Q^{\pi_e}(s_0,a_0)
\right].
\end{equation}

\subsubsection{Weighted Importance Sampling}
Weighted Importance Sampling (WIS) \cite{precup2000eligibility} estimates
policy value by reweighting observed trajectories according to the ratio
between the evaluation policy and the estimated behavior policy. We estimate
the behavior policy using a behavior-cloning model with a small uniform
mixture ($\epsilon=0.05$).

\subsubsection{Weighted Doubly Robust Evaluation}
Weighted Doubly Robust evaluation (WDR) \cite{thomas2016data} combines
importance sampling with the FQE value estimates to reduce variance while
retaining correction from observed trajectories.

\subsection{Implementation Details}
We split the mechanical ventilation episodes into training and testing subsets
in an 80:20 ratio. Within the training set, 5-fold cross-validation was used
to select shared reward and policy-learning hyperparameters. Because this step
tunes parameters that are shared across all state representations,
cross-validation was carried out using the structured-only state
representation, which serves as our baseline. We tuned the Conservative
Q-Learning coefficient $\alpha_{\mathrm{CQL}}$ together with the reward-penalty
coefficients $\alpha$ and $\beta$, which balance improving oxygenation
(measured by per-step changes in SpO$_2$) against discouraging aggressive
ventilator settings. Aggressive actions were defined as 
Vt$_{\mathrm{set}}$ index $\geq 6$ or FiO$_2$ index $\geq 3$,
corresponding to the higher action indices in our discretized action space. The binning intervals used to construct the action space are reported in
Table~\ref{tab:action_binning_interval}.
\begin{table}[t]
\centering
%\small
\begin{tabular}{c c c c}
\toprule
Bin & Vt$_{\mathrm{set}}$ & PEEP & FiO$_2$ (\%) \\
\midrule
1 & $[0.00, 3.94)$   & $[0.00, 7.04)$   & $[0, 38.13)$ \\
2 & $[3.94, 5.36)$   & $[7.04, 11.28)$  & $[38.13, 51.73)$ \\
3 & $[5.36, 6.54)$   & $[11.28, 16.47)$ & $[51.73, 62.12)$ \\
4 & $[6.54, 7.74)$   & $[16.47, \infty)$ & $[62.12, 68.12)$ \\
5 & $[7.74, 9.12)$   & --                & $[68.12, 76.82)$ \\
6 & $[9.12, 11.11)$  & --                & $[76.82, 90.94)$ \\
7 & $[11.11, \infty)$ & --                & $[90.94, 100.00]$ \\
\bottomrule
\end{tabular}
\caption{Action space binning intervals.}
\label{tab:action_binning_interval}
\end{table}

For each candidate setting, we trained a CQL policy on the four training folds
and evaluated it on the validation fold using model-based rollouts. Each
rollout was initialized from the observed initial state of an ICU stay, and at
each subsequent step the learned CQL policy selected an action while the
learned transition model predicted the next state. The SpO$_2$ improvement for
ICU stay $i$ was the mean per-step change in inverse-transformed SpO$_2$ along
the simulated trajectory,
\[
\textstyle\Delta_i^{\mathrm{SpO}_2}
=
\frac{1}{T_i-1}
\sum_{t=0}^{T_i-2}
\left(
\widehat{\mathrm{SpO}}_{2,i,t+1}
-
\widehat{\mathrm{SpO}}_{2,i,t}
\right),
\]
averaged across stays in the validation fold. 
To characterize the accuracy of the transition model used inside model-based
evaluation, we evaluated it by one-step next-state prediction
on the held-out test set. For next-step SpO$_2$ prediction in physiological
units, the model achieved a mean absolute error of $1.080$ and an ROC-AUC of
$0.903$ for classifying whether SpO$_2$ exceeded $94\%$.
The selected configuration
maximized validation SpO$_2$ improvement subject to a safety constraint on the
rate of aggressive actions, yielding $\alpha^{*}=0.375$, $\beta^{*}=0.75$, and
$\alpha_{\mathrm{CQL}}=0.25$.

For all note-based state representations, we fixed representation
hyperparameters before policy learning. The history window was set to
\(k=18\), matching the maximum number of 4-hour decision points in the
72-hour horizon; the embedding-space method used \(\ell=2\) singular vectors,
and each note-derived component was compressed to \(q=64\) dimensions using
training-set PCA. These values were shared across representations to keep state dimensions comparable.

All final models were trained with fixed hyperparameter settings after
cross-validation. The CQL policies used a discount factor $\gamma=0.99$ and
were trained with d3rlpy library \cite{seno2022d3rlpy}, which uses a two-layer MLP
encoder with $256$ hidden units per layer, the Adam optimizer with learning
rate $6.25\times 10^{-5}$. The behavior cloning models
used a two-layer fully connected encoder with $[128,128]$ hidden units,
learning rate $10^{-4}$, batch size $512$, weight decay $10^{-4}$, and the
Adam optimizer. The transition model was a two-layer MLP with hidden
dimension $256$, dropout probability $0.1$, learning rate $10^{-3}$, batch
size $256$, weight decay $10^{-4}$, the Adam optimizer, and early stopping on
held-out next-state MSE. These settings were held fixed across state representations. 
\subsection{Numerical Results}
\begin{figure*}[t]
    \centering
    \includegraphics[width=\textwidth]{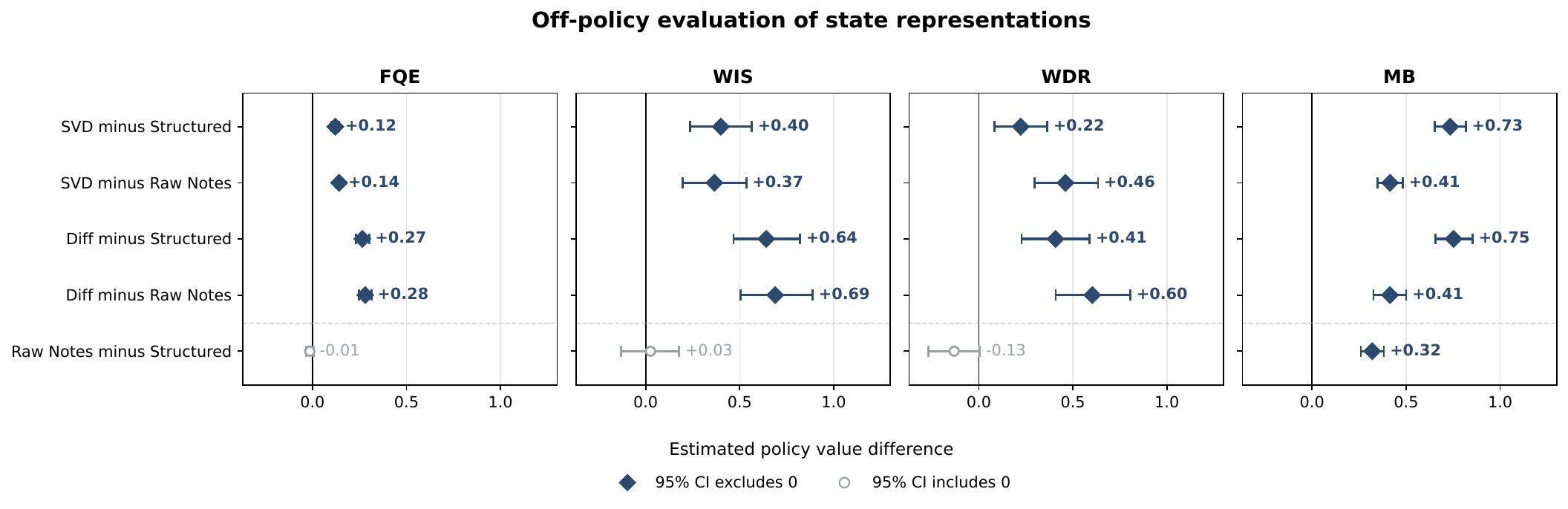}
    \caption{Off-policy evaluation of state representations.
    Paired-bootstrap mean differences in estimated policy value (left row
    label minus right) with $95\%$ CIs. Filled diamonds: CI excludes zero.
    Bottom row (below dashed line): raw-note check.}
    \label{fig:main_ope_comparison}
\end{figure*}

\subsubsection{Aggregate OPE Comparisons Across State Representations}

Figure~\ref{fig:main_ope_comparison} reports paired bootstrap comparisons of
estimated policy value across state representations. Positive values indicate
higher estimated value for the representation on the left than for the baseline on the right. Filled diamonds denote comparisons whose $95\%$ bootstrap
confidence intervals exclude zero, and hollow circles denote comparisons whose
intervals include zero. We evaluate each comparison using four OPE estimators:
FQE, WIS, WDR, and MB.

Both redundancy-aware representations consistently improve over the
structured-only baseline. The embedding-space SVD representation achieves
estimated gains of $+0.12$ (FQE), $+0.40$ (WIS), $+0.22$ (WDR), and $+0.73$
(MB). The sentence-level diff representation achieves larger gains of
$+0.27$, $+0.64$, $+0.41$, and $+0.75$ under the same estimators. All of these
comparisons have confidence intervals that exclude zero.

The redundancy-aware representations also improve over the raw-note baseline.
The SVD representation improves over raw notes by $+0.14$ to $+0.46$, and the
diff representation improves over raw notes by $+0.28$ to $+0.69$, depending
on the estimator. In contrast, the raw-note representation does not show a
consistent improvement over the structured-only state: its confidence intervals include zero under FQE, WIS, and WDR, and only MB gives a positive point estimate. This pattern suggests that the observed gains are not simply due to adding note embeddings, but to representing notes in a way that separates persistent clinical history from newly introduced information.

FQE produces narrower bootstrap intervals than WIS and WDR, reflecting the difference between value function based estimation and trajectory level
importance weighting. Since each estimator has different bias--variance
tradeoffs, we base our conclusions on the consistent improvements observed
across all four OPE methods.
% FQE gives narrower bootstrap intervals than WIS and WDR, which is consistent
% with the fact that FQE averages predictions from a fitted value function,
% whereas WIS and WDR depend on trajectory-level importance weights. At the same
% time, narrower intervals should not be interpreted as evidence that FQE is
% more reliable in general, since fitted value methods may introduce their own
% approximation or extrapolation bias. We therefore base our conclusion on the
% agreement across all four OPE estimators.
\subsubsection{Case Study: Separating Stable Ventilation History}

To better understand how the redundancy-aware note representations affect
policy recommendations, we examine a representative held-out trajectory from
ICU stay 246432. At decision time $t=8$, the clinician selected action 412,
and all learned policies agreed with this choice. At the following decision
time $t=9$, the clinician again selected action 412. The policies learned from
the structured-only, embedding-space SVD, and sentence-level diff
representations also selected action 412, while the policy learned from the
raw-note representation selected action 512.

The notes around this decision are summarized in Table~\ref{tab:case_study_notes}.
At $t=8$, the respiratory care note describes a stable ventilation status,
with stable arterial blood gases and no ventilator changes required. At $t=9$,
the note contains substantial new clinical information, including planned
surgery, possible transfusion, coagulation abnormalities, potassium
replacement, and continued sedation/paralysis. However, the ventilation-related
information remains unchanged: the patient continues to have good arterial
blood gas results on 40\% FiO$_2$ and 5 PEEP, with clear bilateral breath
sounds.

This example illustrates the challenge of using raw note representations for
ventilator control. The raw note embedding may capture newly introduced
information that reflects broader clinical events but is not directly relevant to ventilator adjustment, causing the policy to change its recommendation.
In contrast, the redundancy-aware representations preserve the stable
ventilation context from previous notes while separating newly introduced
information, allowing the learned policies to remain consistent with the
clinician's action. This suggests that redundancy reduction is not only about
removing repeated text, but also about preserving clinically relevant history
for sequential decision-making.

\begin{table}[t]
\centering
%\small
\begin{tabular}{p{0.10\linewidth} p{0.24\linewidth} p{0.56\linewidth}}
\toprule
Time & Note type & Excerpt and interpretation \\
\midrule
$t=8$ &
Respiratory care note &
The patient remained intubated, sedated, and mechanically ventilated. The note
reports stable ABGs, clear and equal bilateral lung sounds, and no ventilator
changes required overnight. \\
\midrule
$t=9$ &
Nursing note &
The note adds new information about planned return to the operating room,
possible transfusion, elevated coagulation measures, potassium replacement,
and continued sedation/paralysis. However, the ventilation-specific content is
stable: good ABG on 40\% FiO$_2$ and 5 PEEP, with clear bilateral breath
sounds. \\
\bottomrule
\end{tabular}
\caption{Clinical notes around decision time $t=9$ for ICU stay 246432.}
\label{tab:case_study_notes}
\end{table}
\section{Conclusion}

In conclusion, we presented a redundancy-aware multimodal state representation
framework for offline RL in mechanical ventilation. Our approach addresses the challenge that longitudinal clinical notes contain substantial temporal redundancy, which can obscure decision-relevant information when raw note embeddings are directly incorporated into RL states. Rather than modifying the RL algorithm, we focus on improving state construction by decomposing clinical notes into history-related and
new-information components before policy learning. We introduce two computationally efficient implementations of this idea: an embedding-space temporal decomposition based on recent note history and a sentence-level diff decomposition that explicitly separates repeated and newly introduced text, which are able to successfully deal with temporally redundant information in clinical notes.

Using MIMIC-III mechanical ventilation trajectories, we show that both redundancy-aware representations improve estimated policy value over structured-only and raw-note multimodal baselines across multiple off-policy evaluation methods. These results demonstrate that clinical notes can provide
useful additional information for offline RL when their temporal redundancy is explicitly modeled. Separating persistent clinical context from newly introduced information helps construct more effective multimodal state
representations for sequential decision-making.

However, these findings should be interpreted within the limitations of
retrospective offline RL. Off-policy evaluation estimates policy performance
from observational data rather than providing direct evidence of clinical
benefit, and learned policies may still be affected by dataset bias,
distribution shift, and limited coverage of state-action pairs. In addition,
our decomposition methods identify textual or representational changes, which may not always correspond to clinically meaningful updates for ventilator control. Finally, this study is based on a single ICU dataset, MIMIC-III, and further validation on more diverse clinical settings is needed before deployment. Future work will investigate combining redundancy-aware
representations with uncertainty estimation, clinical interpretability methods, and prospective evaluation to develop more reliable multimodal RL systems for clinical decision support.

\bibliography{aaai2027}

\end{document}